\documentclass[preprint,12pt]{elsarticle}

\usepackage{amssymb}
\usepackage{amsmath}
\usepackage{xcolor}
\usepackage{adjustbox}
\usepackage[hidelinks]{hyperref}

\journal{Artificial Intelligence in Medicine}

\begin{document}

\begin{frontmatter}



\title{BioNerFlair: biomedical named entity recognition using flair embedding and sequence tagger}


\author[1]{Harsh Patel\corref{cor1}}
\ead{patel.harsh1014@gmail.com}

\address[1]{Department of Computer Science and Engineering, Medi-Caps University, Indore, 453331, India}
\cortext[cor1]{Corresponding author}

\begin{abstract}
\textit{Motivation}: The proliferation of Biomedical research articles has made the task of information retrieval more important than ever. Scientists and Researchers are having difficulty in finding articles that contain information relevant to them. Proper extraction of biomedical entities like Disease, Drug/chem, Species, Gene/protein, can considerably improve the filtering of articles resulting in better extraction of relevant information. Performance on BioNer benchmarks has progressively improved because of progression in transformers-based models like BERT, XLNet, OpenAI, GPT2, etc. These models give excellent results; however, they are computationally expensive and we can achieve better scores for domain specific tasks using other contextual string-based models and LSTM-CRF based sequence tagger.\\
\textit{Results}: We introduce BioNerFlair, a method to train models for biomedical named entity recognition using Flair plus GloVe embeddings and Bidirectional LSTM-CRF based sequence tagger. With almost the same generic architecture widely used for named entity recognition, BioNerFlair outperforms previous state-of-the-art models. I performed experiments on 8 benchmarks datasets for biomedical named entity recognition. Compared to current state-of-the-art models, BioNerFlair achieves the best F1-score of 90.17 beyond 84.72 on the BioCreative II gene mention (BC2GM) corpus, best F1-score of 94.03 beyond 92.36 on the BioCreative IV chemical and drug (BC4CHEMD) corpus, best F1-score of 88.73 beyond 78.58 on the JNLPBA corpus, best F1-score of 91.1 beyond 89.71 on the NCBI disease corpus, best F1-score of 85.48 beyond 78.98 on the Species-800 corpus, while near best results was observed on BC5CDR-chem, BC3CDR-disease, and LINNAEUS corpus.
\end{abstract}

\begin{keyword}
Biomedical named entity recognition \sep Flair embedding \sep GloVe embedding \sep Bidirectional LSTMs \sep Conditional Random Fields \sep Natural Language Processing \sep
\end{keyword}

\end{frontmatter}


\section{Introduction}
\label{}
There is a sharp increase in the number of research papers in the biomedical domain since the pandemic arrived. Scientists around the world are conducting experiments and clinical trials to learn more about the effects of this pandemic on global health and the economy. Because of this, Journals around the world are flooded with biomedical literature and it's getting difficult to find articles that are relevant, robust, and credible. According to different re-ports, over 100,000 papers are already being published for COVID-19 alone. PubMed alone comprises over 30 million citations for biomedical literature. As reports on information about discoveries and insights are added to the already overwhelming amount of literature, the need for advanced computational tools for text mining and information extraction is more important than ever.\par
Recent progress of deep learning techniques in natural language processing (NLP) has led to significant advancements on a wide range of tasks and applications. The domain of biomedical text mining has likewise seen an improvement. The performance in biomedical named entity recognition which automatically extracts entities such as disease, gene/protein, chemicals, species has substantially improved \cite{hong2020dtranner, lee2020biobert}. We can use BioNer for building biomedical knowledge graph. Other NLP domains like entity relation, question answering (QA), depend upon this graph. Thus, improved performance of BioNer can lead to better performance of other complex NLP tasks. Named Entities in biomedical literature have several characteristics that make their extraction from text particularly challenging \cite{zhou2004recognizing}, including the descriptive naming convention (e.g. ‘normal thy-mic epithelial cells’), abbreviations (e.g. ‘IL2’ for ‘Inter-leukin 2’), non-standardized naming convention (e.g. ‘Nace-tylcysteine’, ‘N-acetyl-cysteine’, ‘NAcetylCysteine’, etc.), c-onjunction and disjunction (e.g. ‘91 and 84 kDa proteins’ comprises two entities ‘91 kDa proteins’ and ‘84 kDa proteins’). Traditionally, NER models for biomedical literature perform-ed efficaciously using feature engineering, i.e. carefully selecting features from the text. These features can be linguistic, orthographic, morphological, contextual \cite{campos2012biomedical}. Selecting right features that properly represent target entities requires expert knowledge, lots of trial-error experiments, and is often time consuming whose solution leads to highly specialized models that only works for specialized domains.\par
Models based on convolutional neural networks was proposed to tackle sequence tagging problems \cite{collobert2011natural}. This kind of neural network architecture and learning algorithms reduced the need for domain-specific feature engineering. However, these types of networks could not connect with previous information that could improve performance for Named Entity Recognition. RNN's could capture earlier information through back propagation, but they suffer from the vanishing gradients, exploding gradient problems, and don't handle long-term dependencies well. The gradients carry information for parameter updates. The text data sequences for NER are generally long. For longer sequences, gradients become vanishingly smaller, resulting in no updates of weights \cite{bengio1994learning}. These problems are addressed by a special RNN architecture - Long Short-Term Memory (LSTM), capable of handling long-term dependencies \cite{hochreiter1997long}.\par
The neural architecture - BiLSTM-CRFs produces state-of-the-art performance for NER tasks. This architecture comprises two components: BiLSTM that predict the label by capturing information from the text in both directions and CRF that compute transition compatibility between all possible pairs of labels on neighboring tokens. We now consider this neural architecture standard for sequence labeling problems \cite{lample2016neural}. This kind of architecture generally uses vector representation of words (word embeddings) as input to LSTMs. Word2Vec \cite{mikolov2013distributed}, GloVe \cite{pennington2014glove} are some popular context-independent vector representations of words. Many times, character level features of the text are incorporated into word embeddings layer to improve the performance of NER models \cite{kim2015character}.\par
The use of BiLSTM-CRFs along with certain word embeddings led to significant improvement in the performance of NER models. Researchers starting experimenting with this architecture for Biomedical named entity recognition. Some models used character level embedding along with word embedding pre-trained on a large entity independent corpus (Pub-Med abstracts). These models outperformed earlier state-of-the-art models for BioNER \cite{habibi2017deep, luo2018attention, verwimp2017character}. All the word embeddings used until now were context independ-ent. They cannot address the polysemous and context dependent nature of words. The introduction of contextualized string embeddings such as flair embeddings \cite{akbik2018contextual}, ELMo \cite{peters2018deep} solved this problem. These context-dependent word embeddings when used with BiLSTM-CRFs outperformed all previous models in named entity recognition. Also, transformers based \cite{vaswani2017attention} language representation models like BERT \cite{devlin2018bert} came that achieved state-of-the-art performance in NER. However, applying these NLP methodologies on biomedical literature has limitations because of the different word distribution of general and biomedical corpora. Since recent language representation models are mostly trained in general domain text, they often face problems on biomedical corpora. Most recent state-of-the-art solutions have shown that using a language representation model pre-trained on biomedical corpora (like PubMed abstracts and PMC full-text articles) gives the best results for Biomedical Named Entity Recognition \cite{hong2020dtranner, lee2020biobert}.\par
This paper represents BioNerFlair, a novel architecture for biomedical named entity recognition. BioNerFlair uses contextualized string embeddings Flair (pre-trained on bio-medical domain) along with GloVe embeddings at the token embeddings layer, then a sequence tagger based on BiLSTM-CRFs is used to extract named entities from biomedical literature. I evaluate the performance of BioNerFlair on 8 benchmarks datasets. BioNerFlair outperforms earlier state-of-the-art models on 5 datasets while shows near similar performance of previous models on other 3 datasets.

\section{Materials and methods}
The following sections present a description of the corpora used for evaluation. Furthermore, a technical description of the architecture used along with details of evaluation metrics is given.

\subsection{Datasets}
The statistics of biomedical named entity recognition datasets are listed in Table \ref{tab:1}. BioNerFlair performance is evaluated on eight standard corpora of disease, gene/protein, dru-g/chemical, and species for biomedical Ner: The NCBI \cite{dougan2014ncbi}  and BC5CDR \cite{li2016biocreative} corpus for disease, BC5CDR \cite{li2016biocreative} and BC4CHEMD \cite{krallinger2015chemdner} corpus for drug/chemical, BC2GM \cite{smith2008overview} and JNLPBA \cite{kim2004introduction} corpus for gene/protein, LINNAEUS \cite{gerner2010linnaeus} and Species-800 \cite{pafilis2013species} corpus for species. These datasets are widely used by Biomedical NLP researchers for testing Bio-Ner models. All the datasets are tagged with the IOB tagging scheme. For proper evaluation with other state-of-the-art techniques, the same data split for training, validation, and testing from earlier works \cite{lee2020biobert, wang2019cross} is adopted.

\begin{table}[h]
\caption{Statistics of the biomedical named entity recognition datasets}\label{tab:1}
\begin{center}
\begin{tabular}{lll}
\hline
Datasets     & Entity type  & Number of annotations \\ \hline
NCBI Disease & Disease      & 6881                  \\
BC5CDR       & Disease      & 12694                 \\
BC5CDR       & Drug/Chem.   & 15411                 \\
BC4CHEMD     & Drug/Chem.   & 79824                 \\
BC2GM        & Gene/Protein & 20703                 \\
JNLPBA       & Gene/Protein & 35460                 \\
LINNAEUS     & Species      & 4077                  \\
Species-800  & Species      & 3708                  \\ \hline
\end{tabular}
\end{center}
\begin{center}
\footnotesize{
\textit{Note}: The number of annotations from \cite{habibi2017deep}, \cite{zhu2018clinical}, and \cite{lee2020biobert} is provided. }
\end{center}
\end{table}

\subsection{Model architecture}
\begin{figure}[t]
	\centering
		\includegraphics[scale=.12]{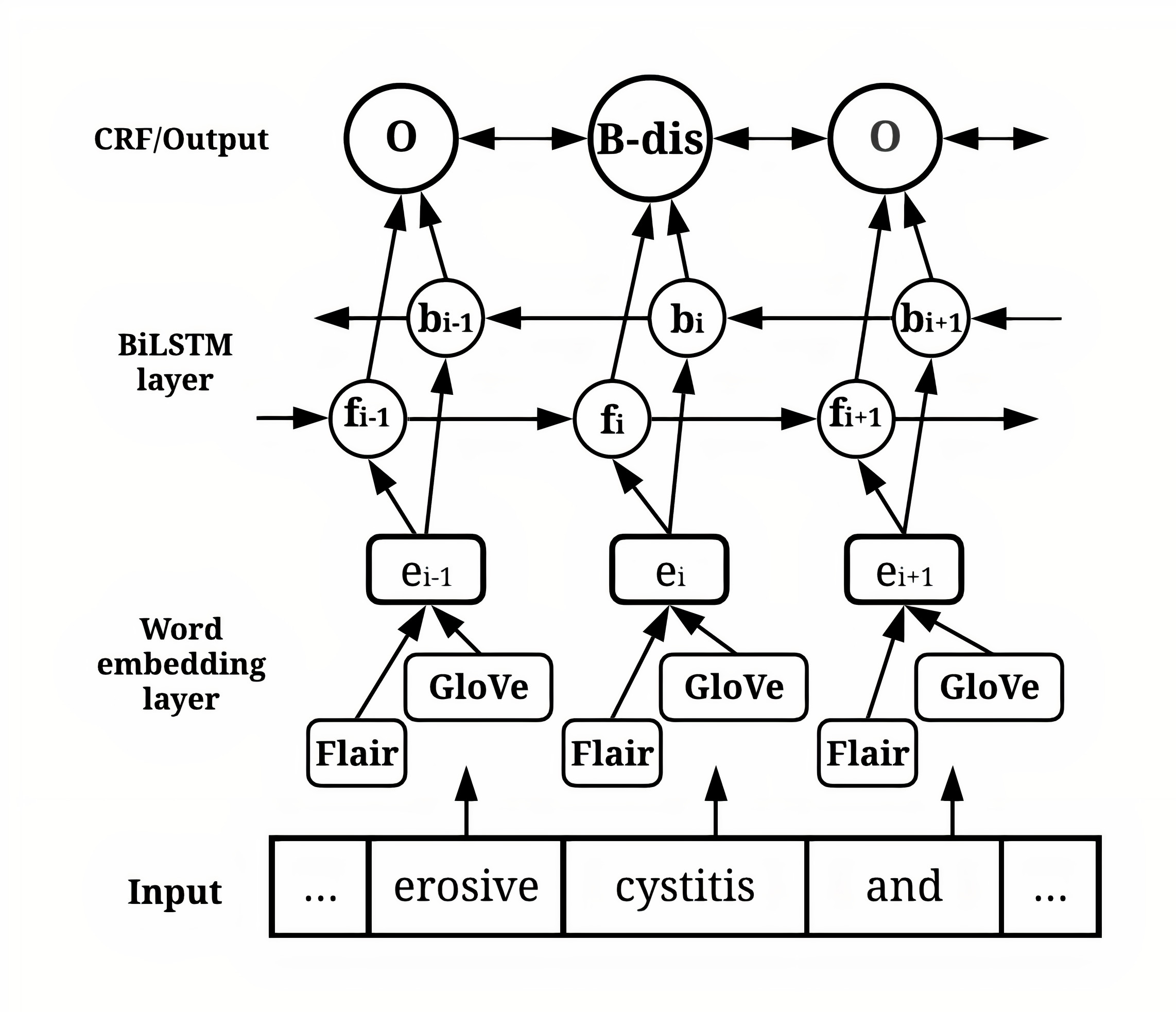}
	\caption{Architecture of BioNerFlair. Flair embedding and GloVe embedding vector representation for a word is computed and concatenated at word embeddings layer. The result is processed by BiLSTM layer and then by CRF layer. The output is the most probable tag sequence, as estimated by CRF.}
	\label{fig:bionerflair}
\end{figure}

BioNerFlair comprises of three layers, namely token embedding layer giving contextualized vector representation of input sequence, passed into vanilla BiLSTM-CRF sequence labeler as depicted in Figure  \ref{fig:bionerflair}, giving state-of-the-art results on BioNer tasks.

\subsubsection{Token embedding layer}
\begin{figure}[h]
	\centering
		\includegraphics[scale=.50]{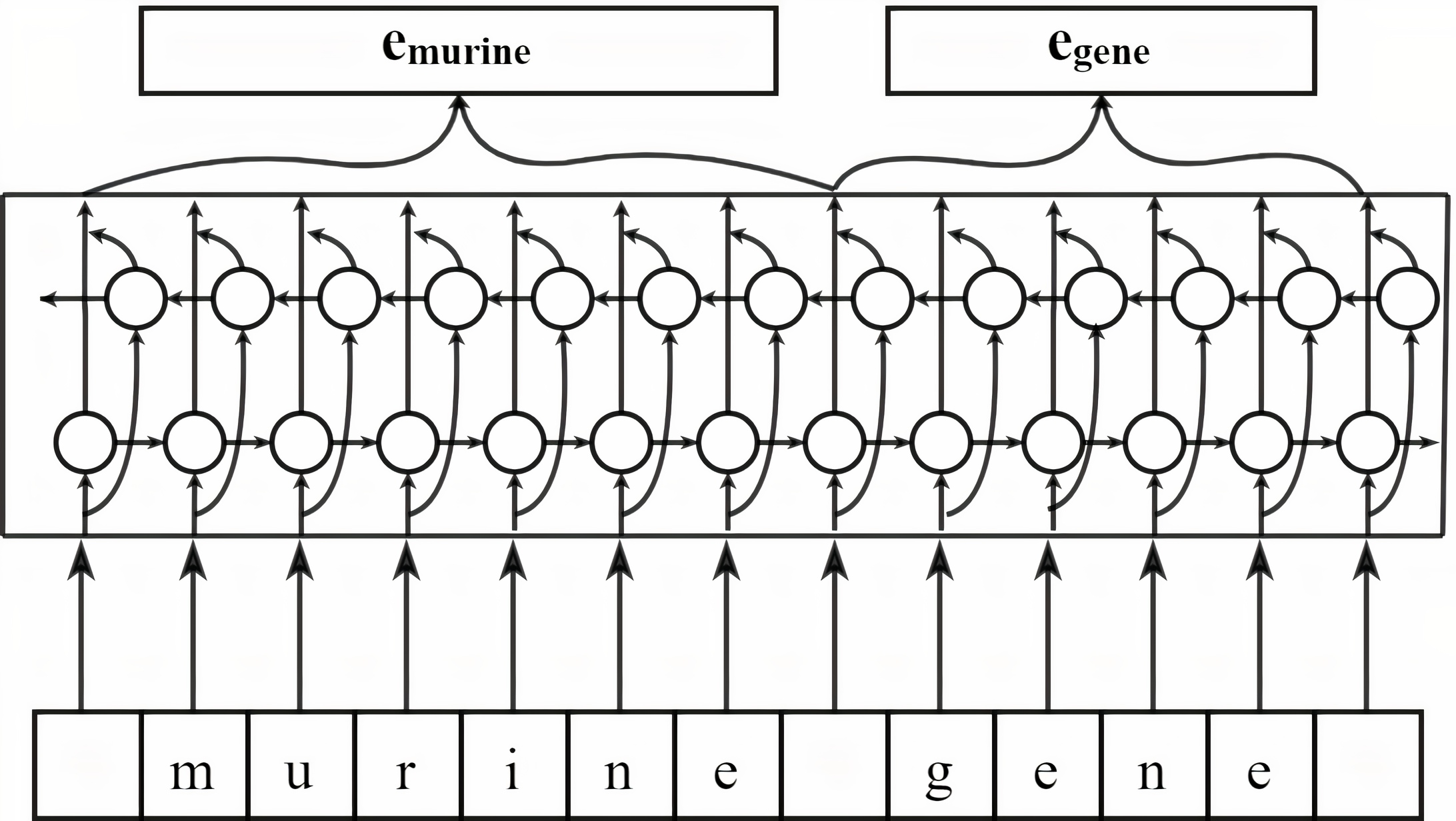}
	\caption{Extraction of flair embeddings in sentential context. It passes the words as a sequence of characters. Output of hidden states are concatenated to form final embedding.}
	\label{fig:flair}
\end{figure}

The token embeddings layer takes as input a sequence of $N$ tokens $(x_{1},x_{2},$ $...,x_{N})$, and outputs a fixed-dimensional vector representation of each token $(e_{1},e_{2},...,e_{N})$. The output here is the concatenation (Equation \ref{equation:1}) of pre-computed GloVe embeddings \cite{pennington2014glove} and contextualized flair embeddings \cite{akbik2018contextual} pre-trained on on roughly 3 million full texts and about 25 million abstracts from the PubMed. Analysis by \cite{akbik2018contextual}, shows that combining flair embeddings with classic world embeddings improves the performance of NER models. In BioNerFlair, GloVe embedding is combined with flair embedding.
\begin{equation}\label{equation:1}
e_i =
\begin{bmatrix}
e^{Flair}_i\\
e^{GloVe}_i
\end{bmatrix}
\end{equation}
Flair embedding is a contextualized character level word embedding that combines the best attributes of different kinds of embeddings. As shown in recent studies \cite{lee2020biobert, dang2018d3ner}, that pre-training models on biomedical corpora significantly improves the performance of BioNer models, this study uses a flair embedding model pre-trained on biomedical data and it seems to capture latent syntactic and semantic similarities. Flair embeddings produce vector representation from hidden states that computes not only on the characters of the word but also the characters of the surrounding context like illustrated in Figure \ref{fig:flair}. Since flair embedding is pre-trained on biomedical corpora and extracts context based on linguistic features at the character level, it handles rare, misspelled, different naming conventions of the words, frequently occurring in biomedical literature very well.

\subsubsection{Bidirectional Long Short-Term Memory (BiLSTM)}
A Long Short Term Memory network (LSTM), is a special kind of RNN introduced by \cite{hochreiter1997long}, explicitly designed to avoid long-term dependency problem. LSTMs does not suffer from vanishing and exploding gradient problems. Unlike RNN, LSTMs can therefore remember information for long periods of time. LSTMs are equipped with memory cells along with an adaptive gating mechanism that regulates the information added or removed from the memory cells. There are three layers in a typical LSTM. A sigmoid layer that decides what information to remove (forget gate), a concatenation of sigmoid and tanh layer that decides what new information to add (input gate), another sigmoid layer that decides the output (output gate). LSTM memory cell is implemented using equations as follows:
\begin{gather}
f_{t} = \sigma(W_{f}\cdot[h_{t-1},x_{t}] + b_{f})\\
i_{t} = \sigma(W_{i}\cdot[h_{t-1},x_{t}] + b_{i})\\
\tilde{C}_{t} = \tanh(W_{c}\cdot[h_{t-1},x_{t}]) + b_{c}\\
C_{t} = (f_{t} \ast C_{t-1}) + (i_{t} \ast \tilde{C}_{t})\\
O_{t} = \sigma(W_{o}\cdot[h_{t-1},x_{t}] + b_{o})\\
h_{t} = O_{t} \ast \tanh(C_{t})
\end{gather}
In the above Equations, $\sigma$ denotes logistic sigmoid function, and i, f, O, and C  are the input gate, forget gate, output gate and cell vectors. In BioNerFlair, the final word embeddings are passed into a BiLSTM network as is seems to capture past features and future features efficiently for a specific time frame \cite{lample2016neural, graves2013speech, huang2015bidirectional}. The bidirectional LSTM network is trained using back-propagation through time \cite{boden2002guide}.

\subsubsection{Conditional Random Fields}
Conditional Random Fields (CRFs) \cite{lafferty2001conditional} is a probabilistic discriminative sequence modeling framework that brings in all the advantages of MEMMs models \cite{ratnaparkhi1996maximum, mccallum2000maximum} while also solving the label bias problem.\par
Given a training dataset $D={(x^1,y^1 ),…,(x^N,y^N )}$ of $N$ data sequences to be labeled $x^i$ and their corresponding label sequences $y^i$, CRFs maximize the log-likelihood of conditional probability of label sequences given their data sequences, that is:
\begin{equation}
L = \sum_{i=1}^{N}{\log(P(y^{i} \vert x^{i}))} - \sum_{k=1}^{K}{\frac{\lambda_{k}^{2}}{2\sigma^{2}}}
\end{equation}

\subsection{Evaluation metrics}
The performance of BioNerFlair is evaluated by training models for each dataset. I used pre-processes versions of BioNer datasets provided by \cite{lee2020biobert}. Also, the same data split is used for training and testing the models. Models are evaluated using precision (P), recall (R), and F1 score metrics on the test corpora. A predicted entity is considered correct if and only if both the entity type and boundary exactly match with annotations in test data. Precision and recall are computed using true positives (TP), false positives (FP), and false negatives (FN). All calculations are done using flair NLP library.
\begin{equation}
P = \frac{TP}{TP + FP}, R = \frac{TP}{TP + FN}, F1 = \frac{2*P*R}{P + R}
\end{equation}

\section{Results and discussion}
\subsection{Experimental setups}
All the models are trained using Flair NLP library, a simple frame-work for state-of-the-art NLP tasks built directly upon PyTorch. I used GPU (12 GB) provided for free by Google Colab to train models. The maximum sequence length was set to 512 to get the best training speed without running out of GPU memory while the mini-batch size for all experiments was set to 32.\par
Model training is started using an initial learning rate of 0.1, patience of 3, and annealing factor of 0.5. A high learning rate of 0.1 works well at starting when using Stochastic Gradient Descent optimizer and is gradually reduced as the model converges. Flair embeddings dropout is set to 0.5. These hyper-parameters are same for all the models. Because of the smaller size of training data and fast GPU, training time of most of the models was less than an hour. However, for the BC4CHEMD dataset, the model could not fit into GPU memory because of which training time increased to around 5 hours.\par
Flair NLP library also comes with Hunflair \cite{weber2020hunflair}, a NER tagger for biomedical text. HunFlair comes with models for genes/proteins, chemicals, diseases, species and cell lines. HunFlair models are trained with multiple datasets at same time due to which it outperforms tools like SciSpacy \cite{neumann2019scispacy} for unseen text but does not give state-of-the-art results on gold standard datasets. In BioNerFlair, I trained models from scratch for each dataset giving results mentioned above. For experiments, I tried to fine tune HunFlair models on target corpus but the model doesn't fit within 12GB of GPU memory.

\begin{table}[ht]
\caption{Test results for biomedical named entity recognition.}\label{table:2}
\begin{adjustbox}{width=1\textwidth}
\small
\begin{tabular}{llllllll}
\hline
Type         & Dataset      & Metrics & SOTA           & DTranNER       & BERT  & BioBERT v1.1   & BioNerFlair    \\ \hline
Disease      & NCBI disease & P       & \underline{88.30}    & 88.21          & 84.12 & 88.22          & \textbf{91.21} \\
             &              & R       & 89.00          & 89.04          & 87.19 & \textbf{91.25} & \underline{91.01}    \\
             &              & F       & 88.60          & 88.62          & 85.63 & \underline{89.71}    & \textbf{91.11} \\
             & BC5CDR       & P       & \textbf{89.61} & 86.75          & 81.97 & 86.47          & \underline{87.88}    \\
             &              & R       & 83.09          & \underline{87.70}    & 82.48 & \textbf{87.84} & 85.73          \\
             &              & F       & 86.23          & \textbf{87.22} & 82.41 & \underline{87.15}    & 86.77          \\
Drug/chem.   & BC5CDR       & P       & \underline{94.26}    & \textbf{94.28} & 90.94 & 93.68          & 91.22          \\
             &              & R       & 92.38          & \textbf{94.04} & 91.38 & \underline{93.26}    & 92.51          \\
             &              & F       & 93.31          & \textbf{94.16} & 91.16 & \underline{93.47}    & 91.85          \\
             & BC4CHEMD     & P       & 92.29          & 91.94          & 91.19 & \underline{92.80}    & \textbf{95.42} \\
             &              & R       & 90.01          & \underline{92.04}    & 88.92 & 91.92          & \textbf{92.72} \\
             &              & F       & 91.14          & 91.99          & 90.04 & \underline{92.36}    & \textbf{94.03} \\
Gene/protein & BC2GM        & P       & 81.81          & 84.21          & 81.17 & \underline{84.32}    & \textbf{89.67} \\
             &              & R       & 81.57          & 84.84          & 82.42 & \underline{85.12}    & \textbf{90.69} \\
             &              & F       & 81.69          & 84.56          & 81.79 & \underline{84.72}    & \textbf{90.17} \\
             & JNLPBA       & P       & \underline{74.43}    & -              & 69.57 & 72.24          & \textbf{86.29} \\
             &              & R       & 83.22          & -              & 81.20 & \underline{83.56}    & \textbf{91.51} \\
             &              & F       & \underline{78.58}    & -              & 74.94 & 77.49          & \textbf{88.73} \\
Species      & LINNAEUS     & P       & \underline{92.80}    & -              & 91.17 & 90.77          & \textbf{97.36} \\
             &              & R       & \textbf{94.29} & -              & 84.30 & \underline{85.83}    & 84.75          \\
             &              & F       & \textbf{93.54} & -              & 87.60 & 88.24          & \underline{90.06}    \\
             & Species-800  & P       & \underline{74.37}    & -              & 69.35 & 72.80          & \textbf{86.83} \\
             &              & R       & \underline{75.96}    & -              & 74.05 & 75.36          & \textbf{84.25} \\
             &              & F       & \underline{74.98}    & -              & 71.63 & 74.06          & \textbf{85.48} \\ \hline
\end{tabular}
\end{adjustbox}
\begin{tiny}
\footnotesize{\textit{Note}:  Marco Precision (P), Recall (R), and F1 (F) scores on each dataset are reported. The best scores are in bold, and the second-best scores are underlined. We list the scores of state-of-the-art (SOTA) models on different datasets as follows:  scores of \cite{xu2019document} on NCBI Disease, scores of \cite{sachan2018effective} on BC2GM,  scores of \cite{lou2017transition} on BC5CDR-disease, scores of \cite{luo2018attention} on BC4CHEMD, scores of \cite{yoon2019collabonet} on BC5CDR-chemical and JNLPBA and scores of \cite{giorgi2018transfer} on LINNAEUS and Species-800. Scores of BioBERT \cite{lee2020biobert} and DTranNER \cite{hong2020dtranner} models are also reported.
}
\end{tiny}
\end{table}

\subsection{Experimental results}
Results of the BioNerFlair method for different datasets are shown in Table \ref{table:2}. The performance of BioNerFlair is compared with other recent state-of-the-art methods. BioNerFlair outperformed state-of-the-art methods on five out of eight datasets while shows near best performance on the remaining three datasets. We can see the biggest improvement in the gene/protein category. BioNerFlair achieves the best F1 score of 90.17 beyond 84.72 on BC2GM corpus and an F1 score of 88.73 beyond 78.58 on JNLPBA corpus. For the species category, BioNerFlair achieves the best F1 score of 85.48 beyond 74.98 on Species-800 corpus, while gets second best score on LINNAEUS corpus. We can notice the same thing for disease and drug/chemical category where BioNerFlair achieves state-of-the-art results of one dataset while getting near best score for other datasets. Even though BioNerFlair does not get best results on BC5CDR corpus for disease and chemical, the results are still competitive when compared with other recent methods and significant improvements can be seen on other datasets.

\subsection{Use of different word embeddings}
In BioNerFlair, I use GloVe embedding and flair embedding at the token embedding layer. Flair NLP library provides the option of Stacked embedding, which allows us to combine different embeddings together. Flair supports classic word embeddings, character embedding, contextualized word embeddings, pre-trained transformer embedding. Therefore, we can experiment with different pairs of embeddings for sequence labeling tasks. The initial plan for this experiment was to use the concatenation of XLNet \cite{yang2019xlnet}, GloVe embedding, and pooled variant of flair embedding \cite{akbik2019pooled}. However, this combination of embeddings requires lots of GPU memory because of which I used the combination of embeddings mentioned above. If more resources are available, we can possibly further improve the performance of BioNer models.

\section{Conclusion}
In conclusion, this article presents BioNerFlair, a metho-d to train models for biomedical named entity recognition using Flair plus GloVe embeddings and a sequence tagger. This paper shows that using contextualized word embedding pre-trained on biomedical corpora significantly improves the results of BioNer models. I evaluated the performance of BioNerFlair on eight datasets. BioNerFlair achieves state-of-the-art results on five datasets. For future study, I plan to experiment with different contextualized and transformer-based word embeddings to further improve the performance of Biomedical Named Entity recognition models.

\section*{Acknowledgements}
I would like to thank the Department of Computer Science and Engineering, Medi-Caps University for the support. I also thank the anonymous reviewers for their comments and suggestions.

\section*{Funding}
This research did not receive any specific grant from fun-ding agencies in the public, commercial, or not-for-profit sectors.

\section*{Availability and implementation}
Source code and data is available at \href{https://github.com/harshpatel1014/BioNerFlair}{https://github.com/harshpatel1014/-BioNerFlair}

\subsection*{Conflict of interest statement}
Declarations of interest: none

\bibliographystyle{elsarticle-num-names} 
\bibliography{cas-refs}

\end{document}